\documentclass[10pt, a4paper]{article}

\usepackage{lrec-coling2024} % this is the new style

\title{What has LeBenchmark Learnt about French Syntax?}

\name{Zdravko Dugonjić,$^{1,2}$*\thanks{*Work done during an internship at Laboratoire d'Informatique de Grenoble.} Adrien Pupier,$^{3}$ Benjamin Lecouteux,$^{3}$ Maximin Coavoux$^{3}$}

\address{$^1$LASR Lab, Technische Universit\"at Dresden, Germany\\
        $^2$School of Embedded and Composite AI (SECAI)\\
        \texttt{first.last@tu-dresden.de}\\
        $^3$Univ. Grenoble Alpes, CNRS, Grenoble INP, LIG, 38000 Grenoble, France\\
        \texttt{first.last@univ-grenoble-alpes.fr}\\}

\abstract{%%% 150 TO 200 words maximum
The paper reports on a series of experiments aiming at probing LeBenchmark,
a pretrained acoustic model trained on 7k hours of spoken French, for syntactic information.
Pretrained acoustic models are increasingly used for downstream speech tasks such
as automatic speech recognition, speech translation, spoken language understanding or speech parsing.
They are trained on very low level information (the raw speech signal), and do not have explicit
lexical knowledge. Despite that, they obtained reasonable results on tasks that requires higher level linguistic knowledge.
As a result, an emerging question is whether these models encode syntactic information.
We probe each representation layer of LeBenchmark for syntax, using the Orféo treebank, and observe that it has learnt some syntactic information. Our results show that syntactic information is more easily extractable from the middle layers of the network, after which a very sharp decrease is observed.
 \\ \newline  }
\begin{document}

\maketitleabstract

\section{Introduction}
The analysis of large pretrained models have emerged as a Natural Language Processing (NLP) subfield aiming at understanding their inner workings, their strengths and weaknesses, as well as interpreting their predictions.

Probing \citep[see][for a general survey]{belinkov-glass-2019-analysis} consists in assessing whether some properties of a model's textual input can be predicted from the intermediate representations of the model.
Probing has been first proposed under the names `auxiliary prediction task' \citep{adi2017finegrained}, or `diagnostic classifiers' \citep{veldhoen2016diagnostic}, as a way to analyze deep learning systems, and in particular understand whether they implicitly learn some knowledge they were not trained on.
For example, despite being trained on raw texts, the various layers of BERT~\citep{devlin-etal-2019-bert} contain a lot of information about the POS tags of its input \citep{tenney2018what,lin-etal-2019-open,rogers-etal-2020-primer}, that can be extracted with a simple linear classifier.

In this paper, we apply probing to a pretrained \textbf{acoustic} model for French, LeBenchmark \citep{evain2021task}, and focus on assessing whether it has implicitly learned some syntactic knowledge.
LeBenchmark is a Wav2vec2.0-style \citep{baevski2020wav2vec} pretrained acoustic model, trained on the raw speech signal, i.e.\ very low level information.
As a result, probing it for high-level information such as syntax is of important significance: syntactic information may be considered several abstractions away from the raw speech signal.

Specifically, we probe LeBenchmark for 2 tasks: part-of-speech tagging and unlabeled dependency parsing.
We frame both tasks as sequence tagging tasks by reducing dependency parsing as a token-level task: predicting the relative position of the head of a token.
We carry out probing using Orféo \citep{benzitoun2016projet}, a treebank of spontaneous French spoken in realistic interactions.
After probing each of the 24 layers of LeBenchmark, we found that syntactic information is most present in the middle layers of the model, and is much less accessible in the last layers where it seems to almost disappear.

In summary our contributions are as follows:
\begin{itemize}
    \item we carry out a probing study on LeBenchmark for syntactic information. This is to the best of our knowledge the first study of this type 
    (i) on French and more generally on a language that is not English
    (ii) on spontaneous speech (rather than read speech).
    \item we report on a finding: syntax is most extractable from the middle layers of the model and almost disappears in the final layers.
\end{itemize}

\begin{figure}
    \centering
    \includegraphics[width=\columnwidth]{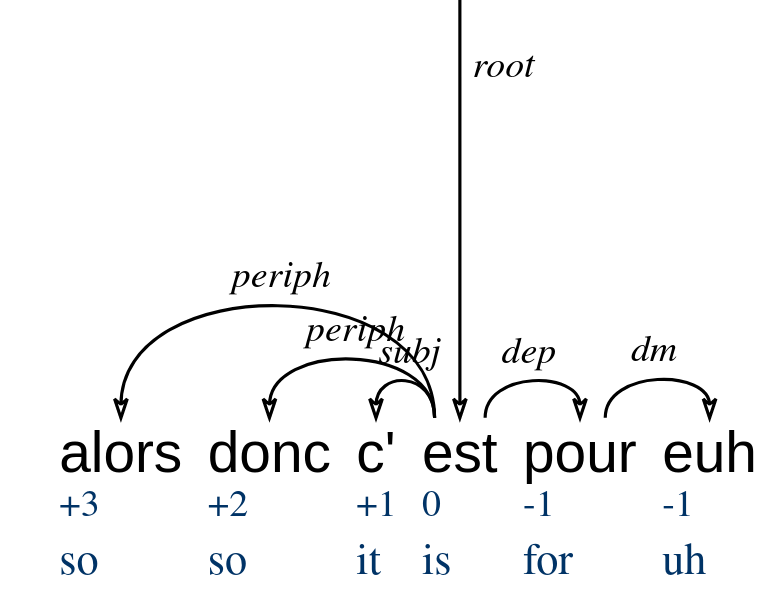}
    \caption{Illustration of the relative head position annotation scheme.}
    \label{fig:tree}
\end{figure}

\section{Related Work}

Self-supervised learning (SSL) consists in acquiring robust representations from extensive unlabeled data (referred to as pretraining) in order to better recognize and understand patterns for other problems (referred to as fine-tuning).
Recent research with a focus on speech data has demonstrated remarkable results in representation learning.
On French \citet{evain2021task} trained and released self-supervised acoustic models called LeBenchmark to address a variety of speech tasks: spoken language understanding, speech-to-text translation, emotion recognition, speech recognition. 
From an acoustic signal, LeBenchmark computes a sequence of vectors, each of which corresponds roughly to a 25ms sound window.
It was pretrained with a constrastive learning objective.
No information about the notion of words or word boundaries is used during the pretraining.

The interest of the community in understanding what information was captured by these models grew recently.
In particular, \citet{canhear} probe a pretrained acoustic model for English, but focus on audio features, the only syntactic features being the depth of the syntactic tree, and the number of occurrences of some parts of speech, where both of these are predicted from the whole sentence representation.
In contrast, we focus on token-level syntactic information.
\citet{shen2023wave} is the study closest to our own: they probe English acoustic models for syntactic information (tree depth, correlations between the continuous representations of the model and the discrete tree representations). 
They found that syntax is best represented in the middle layers of networks, and more obviously in models with larger parameter sets.
Unlike this work, we focus on French, \textit{spontaneous} speech (rather than read speech), and use a different methodology: we focus on simple linear probes.

\section{Data}

We use the Orféo treebank \citeplanguageresource{benzitoun2016projet}, a corpus of spoken French annotated in dependency trees, and distributed with audio recordings.\footnote{\url{https://www.ortolang.fr/market/corpora/cefc-orfeo}}
The Orféo treebank is an aggregation of multiple spoken French corpora, namely CFPP \citeplanguageresource{cfpp2000}, Clapi \citeplanguageresource{11403/clapi/v1}, TCOF \citeplanguageresource{11403/tcof/v2.1}, OFROM \citeplanguageresource{OFROM}, Fleuron \citeplanguageresource{Fleuron}, French Oral Narrative \citeplanguageresource{FON}, c-oral-rom \citeplanguageresource{Coralrom}, 
\textit{Corpus de référence du Français parlé} \citeplanguageresource{delic:halshs-01388193}, Valibel \citeplanguageresource{francard2009corpus}, TUFS \citeplanguageresource{kawaguchi}, a professional meetings corpus \citeplanguageresource{husianycia:tel-01749085}, as well as an unpublished corpus provided by Orféo's designers.
Most of the subcorpora contain French spoken in spontaneous interactions, except for French Oral Narrative
that consists of stories read by narrators.

The total length of recordings is around 196 hours, among which 9 hours have gold syntactic annotations.
The rest of the corpus was annotated with good quality silver syntactic trees \citep{Orfeo-annotation-auto}.
The corpus has around 3.5 million tokens (among which 170k have gold syntactic information.
Finally, the corpus is provided with token timecodes automatically predicted by a forced alignment system. In other words, we have the start and end time of each token.
We discard all sentences that contain obvious timecode mistakes (e.g.\ when the start time of a token is higher than its end time) or annotation mistakes (e.g.\ POS tags that are not listed in the official documentation and correpond to typos in the manual annotation process).

\section{Probing Tasks}
We use two different tasks to assess the amount of syntactic information contained in the pre-trained representations: part-of-speech tagging and unlabelled dependency parsing.
We cast both tasks as word-level prediction tasks.
Though dependency parsing is not typically addressed with sequence tagging methods, recent approaches have shown that it is a viable method \cite{strzyz-etal-2019-viable}.
In our case, addressing both tasks as word-level prediction tasks
has the advantages of keeping the probes and the decoding algorithms fairly simple.

\subsection{POS tagging}
The part of speech tagging task is a classification task where the model classifies each word representation into 20 different parts of speech. 
The tagset is described by \citet{benzitoun2016projet}, It is somewhat finer-grained than the Universal Dependency \citeplanguageresource{nivre-etal-2020-universal} tagset.
The speech transcriptions in the Orféo treebank do not contain punctuation characters. Hence, the tagset does not include punctuation tags.

\subsection{Unlabeled dependency parsing}
In order to cast dependency parsing as a token classification task, we use a simple way of encoding a dependency tree into token labels: relative position encoding \citep{strzyz-etal-2019-viable}.
With this encoding method, the label of a token is an integer corresponding to the relative position of its head word.
This label is computed subtracting the index of the current word to the index of the head word.
In the sentence illustrated in Figure~\ref{fig:tree} : ``Alors donc c' est pour euh"\footnote{Sentence from CEFC-Orféo id: \texttt{cefc-cfpb-1000-5-1}.} (``So so it is for uh"), the root of the sentence is ``est'' and is annotated with the relative label 0, the head of ``Alors'' is ``est'' and thus is annotated with the label +3 (``Alors" is the first word, ``est'' is the fourth one, $4-1=3$).

\begin{figure}
    \centering
    \includegraphics[width=\columnwidth]{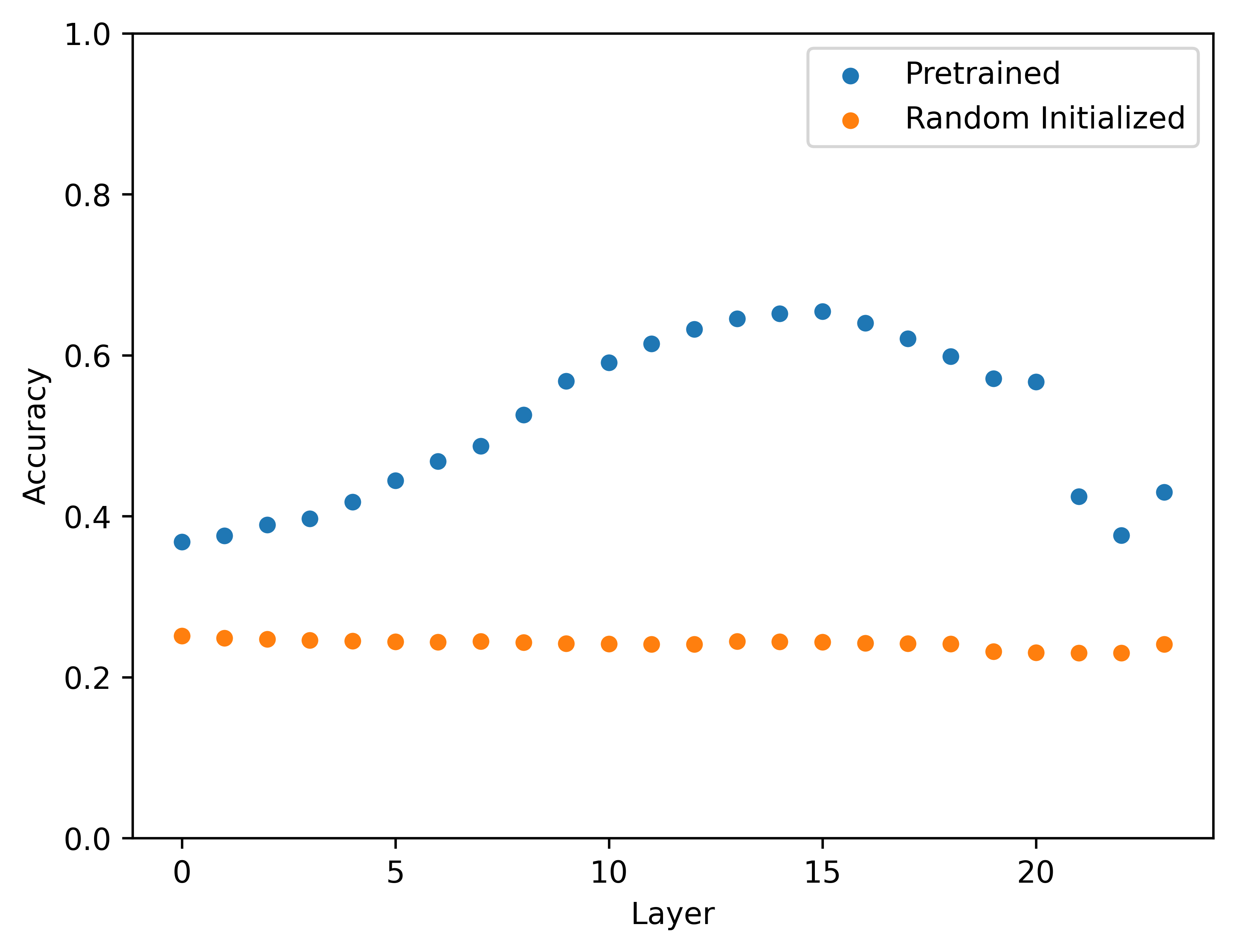}
    \caption{POS tagging task accuracy per layer.}
    \label{fig:pos_accuracy}
\end{figure}

\begin{figure}
    \centering
    \includegraphics[width=\columnwidth]{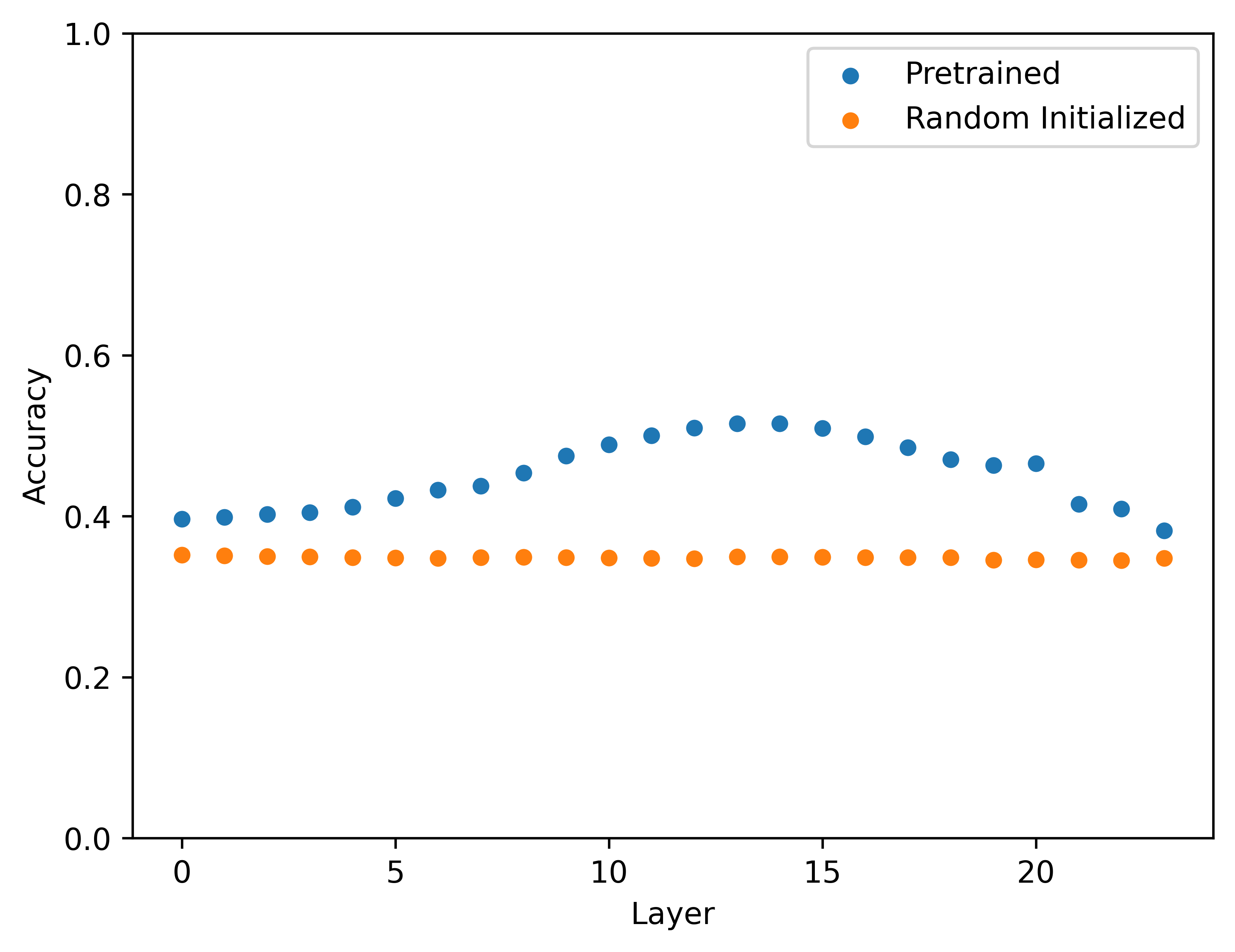}
    \caption{Relative head distance prediction task accuracy (UAS) per layer.}
    \label{fig:rel_accuracy}
\end{figure}

\begin{figure*}
    \centering
    \includegraphics[width=\textwidth]{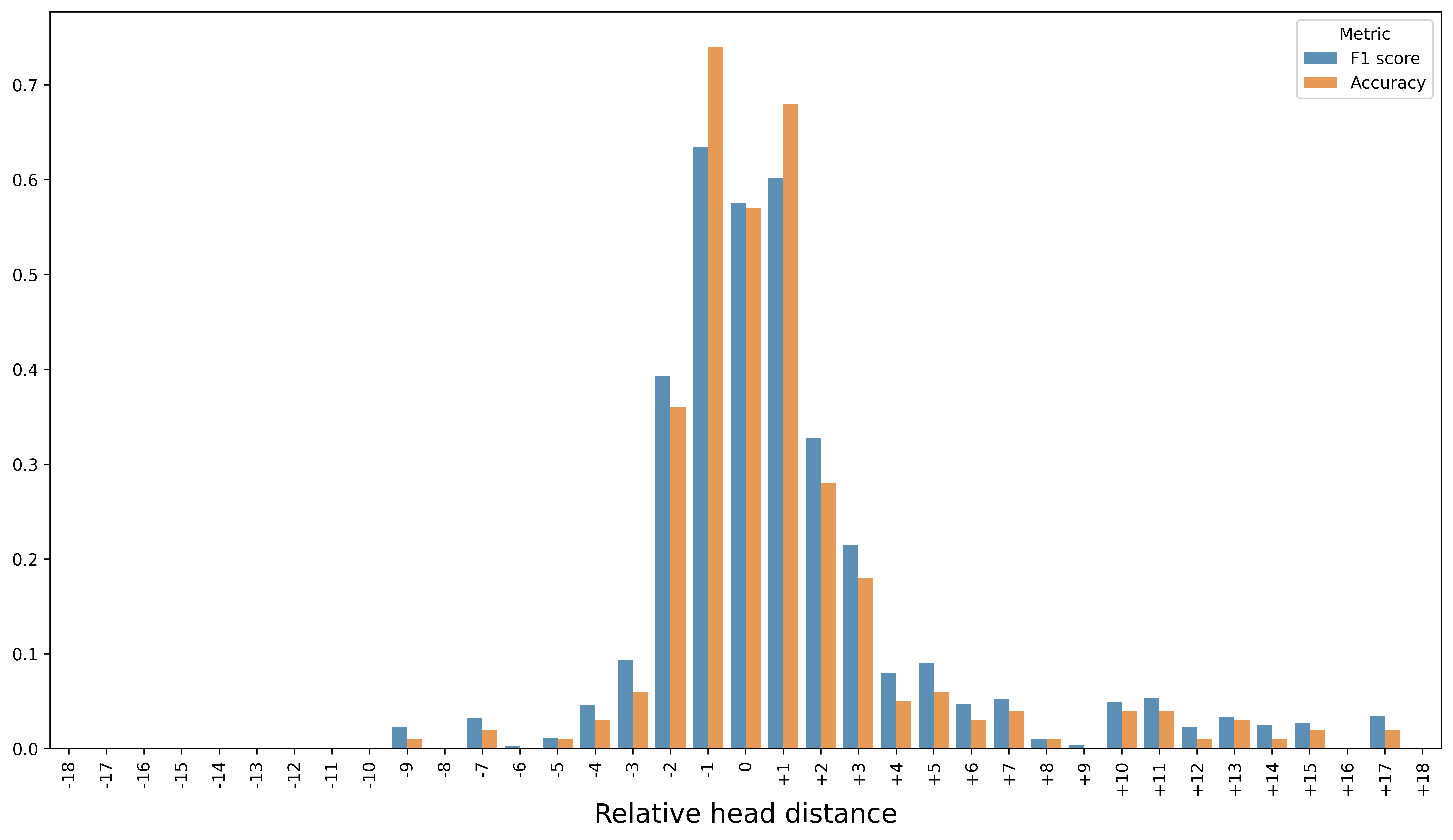}
    \caption{Per-category evaluation of the best layer (layer~14) on the relative head distance prediction task with two evaluation metrics: accuracy (UAS) and F-score.}
    \label{fig:rel_14_f}
\end{figure*}

\section{Experiments}

This section presents the probes we use, outline experimental settings and discuss the experiment results.

\subsection{Probes}

Each layer of the pretrained acoustic model provides a vector representation for each acoustic \textit{frame}. However, both probing tasks require \textit{word} representations.
In order to construct them, we rely on the \textit{start} and \textit{end} timecodes available for each token in the treebank (they were obtained automatically through forced alignement, and sometimes contain mistakes).
We run the pretrained model (whose weights are frozen) on the speech signal, select the vector representations at a specific layer (1-24) and extract, for each token the list of corresponding frames (i.e.\ 1024-coefficient feature vectors) that are within the time span of the token.
Then we aggregate the frames of a token into a single vector with a mean pooling operation.

Finally, we feed token representations to a simple Softmax classifier (one linear projection with bias followed by the Softmax activation).
We repeat the process for each of the 24 layers of the pretrained model in order to probe each of them independently and analyse the dynamics of information across layers.

\subsection{Experimental details}
The self-supervised speech model that we probed in this study is LeBenchmark large\footnote{\url{https://huggingface.co/LeBenchmark/wav2vec2-FR-7K-large}} \citep{evain2021task} based on wav2vec2 \citep{baevski2020wav2vec} architecture,  pre-trained on French datasets containing 7k hours of spontaneous, read, and broadcast speech.

We implemented the probing classifiers in Python, relying on PyTorch \citep{paszke2019pytorch} and HuggingFace \citep{wolf-etal-2020-transformers} libraries. All experiments and preprocessing steps are run on an Nvidia GeForce GTX 980 GPU.

We measured the performance of our probes in both tasks with the accuracy metric, which is a widespread measure in the probing literature. For the relative head distance prediction, the accuracy measure corresponds to the unlabeled attachment score (UAS), a metric classically used in dependency parsing.

During the training, we used a batch size of~1024. The classifier was trained using an early stopping mechanism, which stops the training process if there is no accuracy increase larger than~0.0001 on the validation set over a span of~10 epochs.

For the POS tagging task, we use a learning rate of~0.005 and for the relative head distance prediction task we use a learning rate of~0.001, based on preliminary experiments. For both tasks, we optimize models with \textit{stochastic gradient descent} (SGD) with Nesterov accelerator and momentum~0.99. 
The target optimization function is the negative log-likelihood of the gold labels. 
We use a a random 80\%/10\%/10\% ratio split to construct the training, development and test sets.

\paragraph{Baseline} As a control experiment, we use a baseline that consists of the exact same architecture as the wav2vec2 large model, but where trained parameters are replaced by randomly reinitialized parameters, a classical baseline in the probing literature \citep{arps-etal-2022-probing}.

\subsection{Results}

We present the accuracy of the probe on POS tagging in Figure~\ref{fig:pos_accuracy}, while taking as input each of the~24 transformer layers of the pretrained acoustic model.
The accuracy ranges from~36.8\% for the first layer (layer~0) to~65.5\% for layer~15.
The accuracy decreases after layer~15 and plummets for the last three layers.
In contrast, the random initialisation baseline is about constant across all layers with an average accuracy of 24\%, confirming that LeBenchmark encodes indeed POS information.

We observe a similar behavior on the unlabeled dependency parsing task as shown in Figure~\ref{fig:rel_accuracy}. The accuracy of the classifier reaches the peak of 52\% at the 14th layer and then degrades towards the ends. The average random initialized baseline accuracy is 35\%. The overall accuracy of the classifier on pre-trained representations is lower than in the POS tagging task, which is expected since the task is harder.

Our results are in line with those of \citet{shen2023wave}, who observed a similar pattern on English (using different models, data and methods).
Compared to the literature on probing BERT models for syntax, a remarkable difference is that for both tasks, the accuracy for the final layers plummets to match the accuracy of the first layers (and even that of the baseline in the unlabeled dependency parsing task).
In contrast, \citet{hewitt-manning-2019-structural} also observe that the middle layers of BERT encode the most syntactic information, but the decrease after the middle layers is not as sharp as what we observe for LeBenchmark.

We now consider the results on the unlabeled dependency parsing task using the best layer as input (layer~14) and present its results in terms of accuracy and F$_1$ score broken down by label in Figure~\ref{fig:rel_14_f}. 
As expected, we observe that the model fares better on the most frequent classes, i.e.\ $-1$, $+1$, and~$0$ ($0$ is the label for the root token of an utterance), with F$_1$ scores above 55\%.
Overall, we conclude that the model better encodes \textbf{local} syntactic information.
However, it still has non-zero F$_1$ scores even on longer distance dependency, despite the corpus having no punctuation marks (whose attachement would be more easily predictable).

\section{Conclusion}
We have presented a series of experiments aiming at probing each layer of a French pretrained acoustic model (LeBenchmark) for 2 types of syntactic information: parts of speech, and unlabeled dependency arcs.
We show that the wav2vec2 architecture encode some information about French syntax, in particular local attachments, despite having been pretrained only on raw speech signals.
We show that the middle layers of the model are those from which syntactic information is the more easily extracted, a result in line with recently published results on English \citep{shen2023wave}.
Finally, the accuracy pattern across layers exhibits a sharp decrease in the last layers, a pattern that is not observed in BERT-style text-trained models (where the decrease is much softer).

\section{Limitations}
We acknowledge two limitations of our study. First, due to time constraints we performed all experiments with a single pretrained model, whereas using another model (LeBenchmark base 7k or a multilingual model) would have strengthen the robustness of our findings.
Secondly, a limitation of the type of probe that we use is that the target information might be present in the representations but not extractable with a simple linear classifier.

\section{Ethical Considerations}

To the best of our knowledge, we do not see any potential ethical limitations of our work.

\section{Acknowledgements}

This work is part of the PROPICTO project (French
acronym standing for PRojection du langage
Oral vers des unités PICTOgraphiques), funded by the Swiss National Science Foundation (N°197864) and the French National Research Agency (ANR-20-CE93-0005). 
MC gratefully acknowledges the support of the French National Research Agency (grant ANR-23-CE23-0017-01).
ZD gratefully acknowledges the support of BMBF in DAAD grant 57616814 (\url{https://secai.org/}).

\vfill\eject

\section{Bibliographical References}
\label{sec:reference}

\bibliographystyle{lrec-coling2024-natbib}
\bibliography{lrec-coling2024-example}

\begin{thebibliography}{13}
\expandafter\ifx\csname natexlab\endcsname\relax\def\natexlab#1{#1}\fi

\bibitem[{André(2016)}]{Fleuron}
Virginie André. 2016.
\newblock \href {https://fleuron.atilf.fr/index.php?lg=fr} {Fleuron: Français
  langue Étrangère universitaire–ressources et outils numériques}.

\bibitem[{ATILF(2020)}]{11403/tcof/v2.1}
ATILF. 2020.
\newblock \href {https://hdl.handle.net/11403/tcof/v2.1} {Tcof : Traitement de
  corpus oraux en fran\c{c}ais}.
\newblock {ORTOLANG} ({Open} {Resources} {and} {TOols} {for} {LANGuage})
  \textendash www.ortolang.fr.

\bibitem[{Avanzi et~al.(2012-2020)Avanzi, Béguelin, Corminboeuf, Diémoz, and
  Johnsen}]{OFROM}
Mathieu Avanzi, Marie-José Béguelin, Gilles Corminboeuf, Federica Diémoz,
  and Laure~Anne Johnsen. 2012-2020.
\newblock \href {www.unine.ch/ofrom} {Corpus ofrom – corpus oral de français
  de suisse romande}.
\newblock Université de Neuchâtel.

\bibitem[{Benzitoun et~al.(2016)Benzitoun, Debaisieux, and
  Deulofeu}]{benzitoun2016projet}
Christophe Benzitoun, Jeanne-Marie Debaisieux, and Henri-Jos{\'e} Deulofeu.
  2016.
\newblock Le projet {ORF\'EO}: un corpus d’{\'e}tude pour le fran{\c{c}}ais
  contemporain.
\newblock \emph{Corpus}, (15).

\bibitem[{Carruthers(2013)}]{FON}
Janice Carruthers. 2013.
\newblock French oral narrative corpus.
\newblock Commissioning Body / Publisher: Oxford Text Archive.

\bibitem[{CLESTHIA(2018)}]{cfpp2000}
CLESTHIA. 2018.
\newblock \href {https://hdl.handle.net/11403/cfpp2000/v1} {Cfpp2000}.
\newblock {ORTOLANG} ({Open} {Resources} {and} {TOols} {for} {LANGuage})
  \textendash www.ortolang.fr.

\bibitem[{Cresti et~al.(2004)Cresti, do~Nascimento, Sandoval, Veronis, Martin,
  and Choukri}]{Coralrom}
Emanuela Cresti, Fernanda~Bacelar do~Nascimento, Antonio~Moreno Sandoval, Jean
  Veronis, Philippe Martin, and Khalid Choukri. 2004.
\newblock \href {http://www.lrec-conf.org/proceedings/lrec2004/pdf/357.pdf}
  {The {C}-{ORAL}-{ROM} {CORPUS}. a multilingual resource of spontaneous speech
  for {R}omance languages}.
\newblock In \emph{Proceedings of the Fourth International Conference on
  Language Resources and Evaluation ({LREC}{'}04)}, Lisbon, Portugal. European
  Language Resources Association (ELRA).

\bibitem[{DELIC et~al.(2004)DELIC, Teston-Bonnard, and
  V{\'e}ronis}]{delic:halshs-01388193}
Equipe DELIC, Sandra Teston-Bonnard, and Jean V{\'e}ronis. 2004.
\newblock \href {https://halshs.archives-ouvertes.fr/halshs-01388193}
  {Pr{\'e}sentation du corpus de r{\'e}f{\'e}rence du fran{\c c}ais parl{\'e}}.
\newblock \emph{Recherches sur le fran{\c c}ais parl{\'e}}, 18:11--42.
\newblock Equipe DELIC.

\bibitem[{Francard et~al.(2009)Francard, Hambye, Simon, and
  Dister}]{francard2009corpus}
Michel Francard, Philippe Hambye, Anne-Catherine Simon, and Anne Dister. 2009.
\newblock Du corpus {\`a} la banque de donn{\'e}es.: Du son, des textes et des
  m{\'e}tadonn{\'e}es. l'{\'e}volution de banque de donn{\'e}es textuelles
  orales valibel (1989-2009).
\newblock \emph{Cahiers de l'Institut de linguistique de Louvain-CILL},
  33(2):113.

\bibitem[{Husianycia(2011)}]{husianycia:tel-01749085}
Magali Husianycia. 2011.
\newblock \href {https://hal.univ-lorraine.fr/tel-01749085}
  {\emph{{Caract{\'e}risation de types de discours dans des situations de
  travail}}}.
\newblock Theses, {Universit{\'e} Nancy 2}.

\bibitem[{ICAR(2017)}]{11403/clapi/v1}
ICAR. 2017.
\newblock \href {https://hdl.handle.net/11403/clapi/v1} {Clapi}.
\newblock {ORTOLANG} ({Open} {Resources} {and} {TOols} {for} {LANGuage})
  \textendash www.ortolang.fr.

\bibitem[{Kawaguchi et~al.(2006)Kawaguchi, Zaima, and Takagaki}]{kawaguchi}
Yuji Kawaguchi, Susumu Zaima, and Toshihiro Takagaki, editors. 2006.
\newblock \href {https://www.jbe-platform.com/content/books/9789027292766}
  {\emph{Spoken Language Corpus and Linguistic Informatics}}.
\newblock John Benjamins.

\bibitem[{Nivre et~al.(2020)Nivre, de~Marneffe, Ginter, Haji{\v{c}}, Manning,
  Pyysalo, Schuster, Tyers, and Zeman}]{nivre-etal-2020-universal}
Joakim Nivre, Marie-Catherine de~Marneffe, Filip Ginter, Jan Haji{\v{c}},
  Christopher~D. Manning, Sampo Pyysalo, Sebastian Schuster, Francis Tyers, and
  Daniel Zeman. 2020.
\newblock \href {https://aclanthology.org/2020.lrec-1.497} {{U}niversal
  {D}ependencies v2: An evergrowing multilingual treebank collection}.
\newblock In \emph{Proceedings of the Twelfth Language Resources and Evaluation
  Conference}, pages 4034--4043, Marseille, France. European Language Resources
  Association.

\end{thebibliography}


\begin{thebibliography}{18}
\expandafter\ifx\csname natexlab\endcsname\relax\def\natexlab#1{#1}\fi

\bibitem[{Adi et~al.(2017)Adi, Kermany, Belinkov, Lavi, and
  Goldberg}]{adi2017finegrained}
Yossi Adi, Einat Kermany, Yonatan Belinkov, Ofer Lavi, and Yoav Goldberg. 2017.
\newblock \href {https://openreview.net/forum?id=BJh6Ztuxl} {Fine-grained
  analysis of sentence embeddings using auxiliary prediction tasks}.
\newblock In \emph{International Conference on Learning Representations}.

\bibitem[{Arps et~al.(2022)Arps, Samih, Kallmeyer, and
  Sajjad}]{arps-etal-2022-probing}
David Arps, Younes Samih, Laura Kallmeyer, and Hassan Sajjad. 2022.
\newblock \href {https://aclanthology.org/2022.findings-emnlp.502} {Probing for
  constituency structure in neural language models}.
\newblock In \emph{Findings of the Association for Computational Linguistics:
  EMNLP 2022}, pages 6738--6757, Abu Dhabi, United Arab Emirates. Association
  for Computational Linguistics.

\bibitem[{Baevski et~al.(2020)Baevski, Zhou, Mohamed, and
  Auli}]{baevski2020wav2vec}
Alexei Baevski, Yuhao Zhou, Abdelrahman Mohamed, and Michael Auli. 2020.
\newblock wav2vec 2.0: A framework for self-supervised learning of speech
  representations.
\newblock \emph{Advances in neural information processing systems},
  33:12449--12460.

\bibitem[{Belinkov and Glass(2019)}]{belinkov-glass-2019-analysis}
Yonatan Belinkov and James Glass. 2019.
\newblock \href {https://doi.org/10.1162/tacl_a_00254} {Analysis methods in
  neural language processing: A survey}.
\newblock \emph{Transactions of the Association for Computational Linguistics},
  7:49--72.

\bibitem[{Benzitoun et~al.(2016)Benzitoun, Debaisieux, and
  Deulofeu}]{benzitoun2016projet}
Christophe Benzitoun, Jeanne-Marie Debaisieux, and Henri-Jos{\'e} Deulofeu.
  2016.
\newblock Le projet orf{\'e}o: un corpus d’{\'e}tude pour le fran{\c{c}}ais
  contemporain.
\newblock \emph{Corpus}, (15).

\bibitem[{Devlin et~al.(2019)Devlin, Chang, Lee, and
  Toutanova}]{devlin-etal-2019-bert}
Jacob Devlin, Ming-Wei Chang, Kenton Lee, and Kristina Toutanova. 2019.
\newblock \href {https://doi.org/10.18653/v1/N19-1423} {{BERT}: Pre-training of
  deep bidirectional transformers for language understanding}.
\newblock In \emph{Proceedings of the 2019 Conference of the North {A}merican
  Chapter of the Association for Computational Linguistics: Human Language
  Technologies, Volume 1 (Long and Short Papers)}, pages 4171--4186,
  Minneapolis, Minnesota. Association for Computational Linguistics.

\bibitem[{Evain et~al.(2021)Evain, Nguyen, Le, Boito, Mdhaffar, Alisamir, Tong,
  Tomashenko, Dinarelli, Parcollet, Allauzen, Estève, Lecouteux, Portet,
  Rossato, Ringeval, Schwab, and Besacier}]{evain2021task}
Solène Evain, Ha~Nguyen, Hang Le, Marcely~Zanon Boito, Salima Mdhaffar, Sina
  Alisamir, Ziyi Tong, Natalia Tomashenko, Marco Dinarelli, Titouan Parcollet,
  Alexandre Allauzen, Yannick Estève, Benjamin Lecouteux, François Portet,
  Solange Rossato, Fabien Ringeval, Didier Schwab, and Laurent Besacier. 2021.
\newblock \href {https://doi.org/10.21437/Interspeech.2021-556} {{
  LeBenchmark:} a reproducible framework for assessing self-supervised
  representation learning from speech}.
\newblock In \emph{Proc. Interspeech 2021}, pages 1439--1443.

\bibitem[{Hewitt and Manning(2019)}]{hewitt-manning-2019-structural}
John Hewitt and Christopher~D. Manning. 2019.
\newblock \href {https://doi.org/10.18653/v1/N19-1419} {{A} structural probe
  for finding syntax in word representations}.
\newblock In \emph{Proceedings of the 2019 Conference of the North {A}merican
  Chapter of the Association for Computational Linguistics: Human Language
  Technologies, Volume 1 (Long and Short Papers)}, pages 4129--4138,
  Minneapolis, Minnesota. Association for Computational Linguistics.

\bibitem[{Lin et~al.(2019)Lin, Tan, and Frank}]{lin-etal-2019-open}
Yongjie Lin, Yi~Chern Tan, and Robert Frank. 2019.
\newblock \href {https://doi.org/10.18653/v1/W19-4825} {Open sesame: Getting
  inside {BERT}{'}s linguistic knowledge}.
\newblock In \emph{Proceedings of the 2019 ACL Workshop BlackboxNLP: Analyzing
  and Interpreting Neural Networks for NLP}, pages 241--253, Florence, Italy.
  Association for Computational Linguistics.

\bibitem[{Nasr et~al.(2020)Nasr, Dary, Béchet, and
  Fabre}]{Orfeo-annotation-auto}
Alexis Nasr, Franck Dary, Frederic Béchet, and Benoit Fabre. 2020.
\newblock \href {https://doi.org/https://doi.org/10.3917/lang.219.0087}
  {Annotation syntaxique automatique de la partie orale du {ORFÉO}}.
\newblock In \emph{Langages}.

\bibitem[{Paszke et~al.(2019)Paszke, Gross, Massa, Lerer, Bradbury, Chanan,
  Killeen, Lin, Gimelshein, Antiga et~al.}]{paszke2019pytorch}
Adam Paszke, Sam Gross, Francisco Massa, Adam Lerer, James Bradbury, Gregory
  Chanan, Trevor Killeen, Zeming Lin, Natalia Gimelshein, Luca Antiga, et~al.
  2019.
\newblock Pytorch: An imperative style, high-performance deep learning library.
\newblock \emph{Advances in neural information processing systems}, 32.

\bibitem[{Rogers et~al.(2020)Rogers, Kovaleva, and
  Rumshisky}]{rogers-etal-2020-primer}
Anna Rogers, Olga Kovaleva, and Anna Rumshisky. 2020.
\newblock \href {https://doi.org/10.1162/tacl_a_00349} {A primer in
  {BERT}ology: What we know about how {BERT} works}.
\newblock \emph{Transactions of the Association for Computational Linguistics},
  8:842--866.

\bibitem[{Shen et~al.(2023)Shen, Alishahi, Bisazza, and
  Chrupa{\l}a}]{shen2023wave}
Gaofei Shen, Afra Alishahi, Arianna Bisazza, and Grzegorz Chrupa{\l}a. 2023.
\newblock Wave to syntax: Probing spoken language models for syntax.
\newblock \emph{arXiv preprint arXiv:2305.18957}.

\bibitem[{Singla et~al.(2022)Singla, Shah, Chen, and Shah}]{canhear}
Yaman~Kumar Singla, Jui Shah, Changyou Chen, and Rajiv~Ratn Shah. 2022.
\newblock \href {https://doi.org/10.1109/ICDMW58026.2022.00120} {What do audio
  transformers hear? probing their representations for language delivery \&
  structure}.
\newblock In \emph{2022 IEEE International Conference on Data Mining Workshops
  (ICDMW)}, pages 910--925.

\bibitem[{Strzyz et~al.(2019)Strzyz, Vilares, and
  G{\'o}mez-Rodr{\'\i}guez}]{strzyz-etal-2019-viable}
Michalina Strzyz, David Vilares, and Carlos G{\'o}mez-Rodr{\'\i}guez. 2019.
\newblock \href {https://doi.org/10.18653/v1/N19-1077} {Viable dependency
  parsing as sequence labeling}.
\newblock In \emph{Proceedings of the 2019 Conference of the North {A}merican
  Chapter of the Association for Computational Linguistics: Human Language
  Technologies, Volume 1 (Long and Short Papers)}, pages 717--723, Minneapolis,
  Minnesota. Association for Computational Linguistics.

\bibitem[{Tenney et~al.(2019)Tenney, Xia, Chen, Wang, Poliak, McCoy, Kim,
  Durme, Bowman, Das, and Pavlick}]{tenney2018what}
Ian Tenney, Patrick Xia, Berlin Chen, Alex Wang, Adam Poliak, R~Thomas McCoy,
  Najoung Kim, Benjamin~Van Durme, Sam Bowman, Dipanjan Das, and Ellie Pavlick.
  2019.
\newblock \href {https://openreview.net/forum?id=SJzSgnRcKX} {What do you learn
  from context? probing for sentence structure in contextualized word
  representations}.
\newblock In \emph{International Conference on Learning Representations}.

\bibitem[{Veldhoen et~al.(2016)Veldhoen, Hupkes, Zuidema
  et~al.}]{veldhoen2016diagnostic}
Sara Veldhoen, Dieuwke Hupkes, Willem~H Zuidema, et~al. 2016.
\newblock Diagnostic classifiers revealing how neural networks process
  hierarchical structure.
\newblock In \emph{CoCo@ NIPS}, pages 69--77. Barcelona.

\bibitem[{Wolf et~al.(2020)Wolf, Debut, Sanh, Chaumond, Delangue, Moi, Cistac,
  Rault, Louf, Funtowicz, Davison, Shleifer, von Platen, Ma, Jernite, Plu, Xu,
  Le~Scao, Gugger, Drame, Lhoest, and Rush}]{wolf-etal-2020-transformers}
Thomas Wolf, Lysandre Debut, Victor Sanh, Julien Chaumond, Clement Delangue,
  Anthony Moi, Pierric Cistac, Tim Rault, Remi Louf, Morgan Funtowicz, Joe
  Davison, Sam Shleifer, Patrick von Platen, Clara Ma, Yacine Jernite, Julien
  Plu, Canwen Xu, Teven Le~Scao, Sylvain Gugger, Mariama Drame, Quentin Lhoest,
  and Alexander Rush. 2020.
\newblock \href {https://doi.org/10.18653/v1/2020.emnlp-demos.6} {Transformers:
  State-of-the-art natural language processing}.
\newblock In \emph{Proceedings of the 2020 Conference on Empirical Methods in
  Natural Language Processing: System Demonstrations}, pages 38--45, Online.
  Association for Computational Linguistics.

\end{thebibliography}

\section{Language Resource References}
\label{lr:ref}
\bibliographystylelanguageresource{lrec-coling2024-natbib}
\bibliographylanguageresource{languageresource}

\end{document}